%% file: emnlp2020.tex
\documentclass[11pt,a4paper]{article}
\usepackage[hyperref]{emnlp2020}
\usepackage{times}
\usepackage{latexsym}

\usepackage{microtype}
\usepackage{graphicx}
\usepackage{subfigure}
\usepackage{booktabs} 
\usepackage{natbib}
\usepackage{graphicx}
\usepackage{amsmath}
\usepackage{pdfsync}
\usepackage{amscd}
\usepackage{amssymb}
\usepackage{amsthm}
\usepackage{mathabx}
\usepackage{algorithmic}
\usepackage{algorithm}

\input{math_commands.tex}

\usepackage{xcolor}
\usepackage{tikz}
\usetikzlibrary{automata, arrows}
\usepackage{pgfplots}
\usepackage{tikz}
\usetikzlibrary{arrows, positioning, calc}
\definecolor{color1}{RGB}{27,158,119}
\definecolor{color3}{RGB}{217,95,2}
\definecolor{color2}{RGB}{117,112,179}

\usepackage{latexsym}
\usepackage{verbatim}
\usepackage[utf8]{inputenc}
\usepackage{pdfsync}
\usepackage{caption}

\newcommand{\En}{\texttt{En}}
\newcommand{\Fr}{\texttt{Fr}}
\newcommand{\Ro}{\texttt{Ro}}

\newcommand{\De}{\texttt{De}}
\newcommand{\Es}{\texttt{Es}}
\newcommand{\enc}{\textrm{enc}}

\newcommand{\Y}{\texttt{Y}}

\newcommand{\Cs}{\texttt{Cs}}

\usepackage{amsmath}
\usepackage{hyperref}
\usepackage{microtype}

\aclfinalcopy 
\def\aclpaperid{2310} 

\pgfplotsset{every axis/.append style={
        scaled x ticks = false, 
        x tick label style={/pgf/number format/.cd, fixed, fixed zerofill,
                            int detect, 1000 sep={},precision=3}
    }
}
\title{A Multilingual View of Unsupervised Machine Translation}

\date{}
\author{Xavier Garcia\thanks{\quad Work done as part of the Google AI Residency.}\\ \qquad \qquad \qquad \qquad \qquad \qquad \qquad \qquad \qquad \qquad \qquad \qquad \qquad \qquad \qquad \texttt{\{xgarcia,pierreforet,tsellam,aparikh\}@google.com} \\\And Pierre Foret\footnotemark[1] \\\And Thibault Sellam  \\\And Ankur P. Parikh}
\begin{document}
\maketitle
\begin{abstract}
 We present a probabilistic framework for multilingual neural machine translation that encompasses supervised and unsupervised setups, focusing on unsupervised translation. In addition to studying the vanilla case where there is only monolingual data available, we propose a novel setup where one language in the (source, target) pair is not associated with any parallel data, but there may exist auxiliary parallel data that contains the other. This auxiliary data can naturally be utilized in our probabilistic framework via a novel cross-translation loss term. Empirically, we show that our approach results in higher BLEU scores over state-of-the-art unsupervised models on the WMT'14 English-French, WMT'16 English-German, and WMT'16 English-Romanian datasets in most directions.
\end{abstract}

\section{Introduction}

The popularity of neural machine translation systems \cite{kalchbrenner13,sutskever14,bahdanau14,wu2016google} has exploded in recent years. Those systems have obtained state-of-the-art results for a wide collection of language pairs, but they often require large amounts of parallel (source, target) sentence pairs to train~\citep{koehn2017six}, making them impractical for scenarios with resource-poor languages. As a result, there has been interest in \textit{unsupervised machine translation}~\cite{ravi11}, and more recently \textit{unsupervised neural machine translation} (UNMT)~\cite{lample17,artetxe17}, which uses only monolingual source and target corpora for learning.  Unsupervised NMT systems have achieved rapid progress recently~\citep{lample19,artetxe19,ren19,D2GPO}, largely thanks to two key ideas: one-the-fly back-translation (i.e., minimizing round-trip translation inconsistency) \citep{bannard05,sennrich2015improving,he2016dual, artetxe17} and pretrained language models~\citep{lample19,song19}. Despite the difficulty of the problem, those systems have achieved surprisingly strong results.

\begin{figure}[t!]
  \centering
  \subfigure[{\scriptsize Multilingual NMT}]{
\begin{tikzpicture}[>=stealth',shorten >=1pt,auto,node distance=1.4cm]
  \node[state]         (en) [] {$\textbf{Fr}$};
  \node[state]         (fr) [above left of=en]  {$\textbf{En}$};
  \node[state]         (ro) [above right of=en] {$\textbf{Ro}$};

  \path[-]          (en)  edge                 node {} (ro);
  \path[-]          (en)  edge                 node {} (fr);
  \path[<->, color=blue, thick]          (fr)  edge      [bend left, dotted]           node {} (ro);
   \path[-]         (fr)  edge          []       node {} (ro);
\end{tikzpicture}
\label{pic:supervised}}
  \subfigure[{\scriptsize Zero-Shot NMT}]{
\begin{tikzpicture}[>=stealth',shorten >=1pt,auto,node distance=1.4cm]
  \node[state]         (fr) [above left of=en]  {$\textbf{En}$};
  \node[state] (en) [] {$\textbf{Fr}$};
  \node[state]         (ro) [above right of=en] {$\textbf{Ro}$};

  \path[<->, color=blue, thick]          (fr)  edge      [bend left, dotted]           node {} (ro);
  \path[-]          (en)  edge                 node {} (ro);
  \path[-]          (en)  edge                 node {} (fr);
\end{tikzpicture}
\label{pic:zeroshot}
  }
  \\
  \subfigure[{\scriptsize M-UNMT (w/o auxiliary parallel data)}]{
\begin{tikzpicture}[>=stealth',shorten >=1pt,auto,node distance=1.4cm]
  \node[state]         (fr) [above left of=en]  {$\textbf{En}$};
  \node[state] (en) [] {$\textbf{Fr}$};
  \node[state]         (ro) [above right of=en] {$\textbf{Ro}$};

  \path[<->, color=blue, thick]          (fr)  edge      [bend left, dotted]           node {} (ro);
\end{tikzpicture}
\label{pic:unmtnoaux}
  }
  \subfigure[{\scriptsize M-UNMT (w/ auxiliary parallel data)}]{
\begin{tikzpicture}[>=stealth',shorten >=1pt,auto,node distance=1.4cm]\label{pic:unmtwaux}
  \node[state]         (fr) [above left of=en]  {$\textbf{En}$};
  \node[state] (en) [] {$\textbf{Fr}$};
  \node[state]         (ro) [above right of=en] {$\textbf{Ro}$};

  \path[<->, color=blue, thick]          (fr)  edge      [bend left, dotted]           node {} (ro);
  \path[-]          (en)  edge            node {} (fr);
\end{tikzpicture}
  }
 \caption{Different setups for English ($\En$), French ($\Fr$) and Romanian ($\Ro$). The dashed edge indicates the target language pair. Full edges indicate the existence of parallel training data. }
  \label{configs}
\end{figure}
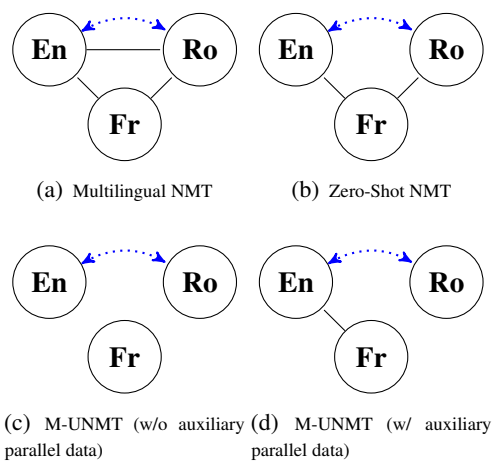



In this work, we investigate \emph{Multilingual UNMT} (M-UNMT), a generalization of the UNMT setup that involves more than two languages. Multilinguality has been explored in the supervised NMT literature, where it has been shown to enable information sharing among related languages. This allows higher resource language pairs (e.g. English--French) to improve performance among lower resource pairs (e.g., English--Romanian)~\citep{johnson16,firat16}. Yet multilingual translation has only received little attention in the unsupervised literature, and
the performance of  preliminary works~\citep{sen19,xu19} is considerably below that of  state-of-the-art bilingual unsupervised systems~\citep{lample19,song19}. Another line of work has studied zero-shot translation in the presence of a ``pivot'' language, e.g., using French-English and English-Romanian corpora to model French-Romanian~\citep{johnson16,arivazhagan19a,gu19,al19}. However, zero-shot translation is not unsupervised since one can perform  two-step supervised translation through the pivot language.

We introduce a novel probabilistic formulation of multilingual translation, which encompasses not only existing supervised and zero-shot setups, but also two variants of Multilingual UNMT: \textbf{(1)} a strict M-UNMT setup in which there is no parallel data for any pair of language, and 
\textbf{(2)} a novel, looser setup where there exists parallel data that contains one language in the (source, target) pair but not the other. 
We illustrate those two variants and contrast them to existing work in Figure~\ref{configs}. As shown in Figures~\ref{configs}(c) and \ref{configs}(d), the defining feature of M-UNMT  is that the (source, target) pair of interest is not connected in the graph, precluding the possibility of any direct or multi-step supervised solution. Leveraging auxiliary parallel data for UNMT as shown in Figure \ref{configs}(d) has not been well studied in the literature. However, this setup may be more realistic than the strictly unsupervised case since it enables the use of high resource languages (e.g. \En) to aid translation into rare languages.



For the strict M-UNMT setup pictured in Figure~\ref{configs}(c), our probabilistic formulation yields a multi-way back-translation objective that is an intuitive generalization of existing work~\citep{artetxe17, lample17, he20}. We provide a rigorous derivation of this objective as an application of the Expectation Maximization algorithm~\citep{dempster77}. Effectively utilizing the auxiliary parallel corpus pictured in Figure~\ref{configs}(d) is less straightforward since the common approaches for UNMT are explicitly designed for the bilingual case. For this setting, we propose two algorithmic contributions. First, we derive a novel \emph{cross-translation} loss term from our probabilistic framework that enforces cross-language pair consistency. Second, we utilize the auxiliary parallel data for \emph{pre-training}, which allows the model to build representations better suited to translation.

Empirically, we evaluate both setups, demonstrating that our approach of leveraging auxiliary parallel data offers quantifiable gains over existing state-of-the-art unsupervised models on 3 language pairs: $\En-\Ro$, $\En-\Fr$, and $\En-\De$.  Finally, we perform a series of ablation studies 
that highlight the impact of the additional data, our additional loss terms, as well as the choice of auxiliary language.

\section{Background and Overview}

\paragraph{Notation:} Before discussing our approach, we introduce some notation. We denote random variables by capital letters $X$, $Y$, $Z$, and their realizations by their corresponding lowercase version $x$, $y$, $z$. We abuse this convention to compactly write objects like the conditional density $p(Y = y |X =x)$ as $p(y|x)$ or the marginalized distributions $p(X=x)$ as $p(x)$, with the understanding that the lowercase variables are connected to their corresponding uppercase random variables. Given a random variable $X$, we write $\E_{x \sim X}$ to mean the expectation with respect to $x$, where $x$ follows the distribution of $X$. We use a similar convention for conditional distributions e.g. we write $\E_{y \sim p(\cdot|x)}$ to denote the expectation of $Y$ conditioned on $X = x$.  Similarly, we write $H(X)$ or $H(p(x))$ to denote the entropy of the random variable $X$ i.e. $H(X) = \E_{x \sim X} [ -\log p(x)]$. We reserve the use of typewriter font for languages e.g. $\texttt{X}$. 

\paragraph{Neural Machine Translation:} In bilingual supervised machine translation we are given a training dataset $\gD_{\textbf{x}, \textbf{y}}$. Each $(x, y) \in \gD_{\textbf{x}, \textbf{y}}$ is a (source, target) pair
consisting of a sentence $x$ in language $\texttt{X}$ and a semantically equivalent sentence $y$ in language $\texttt{Y}$. We train a translation model using maximum likelihood:
\begin{align*}
L_{sup}(\theta) = \sum_{(x,y) \in \gD_{\textbf{x}, \textbf{y}}} \log p_{\theta}(y | x) 
\end{align*}
In neural machine translation, $p_{\theta}(y | x)$ is modelled with the encoder-decoder paradigm where $x$ is encoded into a set of vectors via a neural network $\enc_{\theta}$ and a decoder neural network defines $p_{\theta}(y | \enc_{\theta}(x))$. In this work, we use a transformer~\cite{vaswani17} as the encoder and decoder network. At inference time, computing the most likely target sentence $y$ is intractable since it requires enumerating over all possible sequences, and is thus approximated via beam search.

\paragraph{Unsupervised Machine Translation:} The requirement of a training dataset $\gD_{\textbf{x}, \textbf{y}}$ with source-target pairs can often be prohibitive for rare or low resource languages. Bilingual unsupervised translation attempts to learn $p_{\theta}(y | x)$  using monolingual corpora $\gD_{\vx}$ and $\gD_{\vy}$. For each sentence $x \in \gD_{\vx}$, $\gD_{\vy}$ may not contain an equivalent sentence in $\Y$, and vice versa.

State of the art unsupervised methods typically work as follows. They first perform pre-training and learn an initial set of parameters $\theta$ based on a variety of language modeling or noisy reconstruction objectives~\cite{lample19,lewis19,song19} over $\gD_{\vx}$ and $\gD_{\vy}$. A fine-tuning stage then follows which typically uses back-translation~\citep{sennrich16,lample19,he2016dual} that involves translating $x$ to the target language $\texttt{Y}$, translating it back to a sentence $x'$ in $\texttt{X}$, and penalizing the reconstruction error between $x$ and $x'$.

\paragraph{Overview of our Approach:} 
The following sections describe a probabilistic MT framework that justifies and generalizes the aforementioned approaches.  We first model the case where we have access to several monolingual corpora, pictured in Figure~\ref{pic:unmtnoaux}. We introduce light independence assumptions to make the joint likelihood tractable and derive a lower bound, obtaining a generalization of the back-translation loss. We then extend our model to include the auxiliary parallel data pictured in Figure~\ref{pic:unmtwaux}. We demonstrate the emergence of a \emph{cross-translation} loss term, which binds distinct pairs of languages together. Finally, we present our complete training procedure, based on the EM algorithm. Building upon existing work~\cite{song19}, we introduce a pre-training step that we run before maximizing the likelihood to obtain good representations.

\section{Multilingual Unsupervised Machine Translation}
\label{sec:M-UNMT}

In this section, we formulate our approach for M-UNMT. We restrict ourselves to three languages, but the arguments naturally extend to an arbitrary number of languages. Inspired by the recent style transfer literature~\cite{he20} and some approaches from multilingual supervised machine translation \cite{ren2018triangular}, we introduce a generative model of which the available data can be seen as partially-observed samples. We first investigate the strict unsupervised case, where only monolingual data is available. Our framework naturally leads to an aggregate back-translation loss that generalizes previous work. We then incorporate the auxiliary corpus, introducing a novel cross-translation term. To optimize our loss, we leverage the EM algorithm, giving a rigorous justification for the stop-gradient operation that is usually applied in the UNMT and style transfer literature~\cite{lample19, artetxe19, he20}.

\subsection{M-UNMT - Monolingual Data Only}
\label{subsec:monolingual}

We begin with the assumption that we have three sets of monolingual data, $\mathcal{D}_{\textbf{x}}, \mathcal{D}_{\textbf{y}}, \mathcal{D}_{\textbf{z}}$ for languages $\texttt{X}, \texttt{Y}$ and $\texttt{Z}$ respectively. We take the viewpoint that these datasets form the visible parts of a larger dataset $\mathcal{D}_{\textbf{x},\textbf{y},\textbf{z}}$ of triplets $(x,y,z)$ which are translations of each other. We think of these translations as samples of a triplet $(X,Y,Z)$ of random variables and write the observed data log-likelihood as:
\begin{align*}
\mathcal{L}(\theta) = \mathcal{L}_{\mathcal{D}_{\textbf{x}}} +
\mathcal{L}_{\mathcal{D}_{\textbf{y}}}  + \mathcal{L}_{\mathcal{D}_{\textbf{z}}} 
\end{align*}

Our goal however is to learn a conditional translation model $p_{\theta}$. We thus rewrite the log likelihood as a marginalization over the unobserved variables for each dataset as shown below:
\begin{align}
    \mathcal{L}(\theta) &=  \sum_{x \in \mathcal{D}_x} \log \underset{\underset{\sim (Y, Z)}{(y,z)}}{\E} \, p_{\theta}(x|y,z) \label{eq:first-term} \\ 
    &+ \, \sum_{y \in \mathcal{D}_{\textbf{y}}} \log \underset{\underset{\sim (X, Z)}{(x,z)}}{\E} \, p_{\theta}(y|x,z) \label{eq:second-term} \\ 
    &+ \, \sum_{z \in \mathcal{D}_{\textbf{z}}} \log \underset{\underset{\sim (X, Y)}{(x,y)}}{\E} \, p_{\theta}(z|x,y)
\label{eq:third-term}
\end{align}

Learning a model for $p_{\theta}(x|y,z)$ is not practical since the translation task is to translate $z \rightarrow x$ without access to $y$, or $y \rightarrow x$ without access to $z$. Thus, we make the following structural assumption: given any variable in the triplet $(X,Y,Z)$, the remaining two are independent. We implicitly think of the conditioned variable as detailing the content and the two remaining variables as independent manifestations of this content in the respective languages. Using the fact that $p_{\theta}(x|y,z) = p_{\theta}(x|y) = p_{\theta}(x|z)$ under this assumption, we rewrite the summand in $(1)$ as follows:
$$ \log \underset{\underset{\sim (Y, Z)}{(y,z)}}{\E} p_{\theta}(x|y,z)=\log \underset{\underset{\sim (Y, Z)}{(y,z)}}{\E} \sqrt{p_{\theta}(x | y)p_{\theta}(x | z)}.$$
Next, note that all these expectations in Eq.~\ref{eq:first-term},~\ref{eq:second-term}, and~\ref{eq:third-term} are intractable to compute due to the number of possible sequences in each language. We address this problem through the Expectation Maximization (EM) algorithm~\citep{dempster77}. We first use Jensen's inequality\footnote{This is actually an equality in this case since $\frac{p_{\theta}(x|y,z)}{p_{\theta}(y,z|x)}p(y,z) = p(x)$ and hence the expectant does not actually depend on $y$ or $z$. }:
 \begin{align*}
\log\hspace{-4pt}\underset{\underset{\sim (Y, Z)}{(y,z)}}{\E}&\hspace{-2pt}p_{\theta}(x|y,z) = \log\hspace{-4pt}\underset{\underset{\sim (Y, Z)}{(y,z)}}{\E} \frac{p_{\theta}(x|y,z)}{p_{\theta}(y,z|x)} p_{\theta}(y,z|x) \\
&= \log \underset{\underset{\sim p_{\theta}(y,z|x)}{(y,z)}}{\E} \frac{p_{\theta}(x|y,z)}{p_{\theta}(y,z|x)} p(y,z) \\
&= \underset{\underset{\sim p_{\theta}(y,z|x)}{(y,z)}}{\E}[\log p_{\theta}(x|y,z) + \log p(y,z)]\\ 
&\quad + H(p_{\theta}(y,z|x)) 
 \end{align*}
Since the entropy of a random variable is always non-negative, we can bound the quantity on the right from below as follows:
\begin{align*}
\log \underset{\underset{\sim (Y, Z)}{(y,z)}}{\E} p_{\theta}(x|y,z) &\geq  \underset{\underset{\sim p_{\theta}(y,z|x)}{(y,z)}}{\E}[\log p_{\theta}(x|y,z)] \\
&+ \underset{\underset{\sim p_{\theta}(y,z|x)}{(y,z)}}{\E} [\log p_{\theta}(y,z)] \\
&= \frac{1}{2}\underset{y \sim p_{\theta}(y|x)}{\E} \log p_{\theta}(x|y) \\
&+ \, \frac{1}{2}\underset{z \sim p_{\theta}(z|x)}{\E} \log p_{\theta}(x|z) \\
&+ \underset{\underset{\sim p_{\theta}(y,z|x)}{(y,z)}}{\E} \log p(y,z)
\end{align*}
Applying the above strategy to $(2)$ and $(3)$ and rearranging terms gives us:
\begin{align}
&\gL(\theta) \geq \frac{1}{2} \underset{y \sim p_{\theta}(\cdot | x)}{\E} \log p_{\theta}(x|y) \notag \\
& + \frac{1}{2} \underset{z \sim p_{\theta}(\cdot | x)}{\E} \log p_{\theta}(x|z) + \frac{1}{2} \underset{x \sim p_{\theta}(\cdot | y)}{\E} \log p_{\theta}(y|x) \notag \\
& + \frac{1}{2} \underset{z \sim p_{\theta}(\cdot | y)}{\E} \log p_{\theta}(y|z) + \frac{1}{2} \underset{y \sim p_{\theta}(\cdot | z)}{\E} \log p_{\theta}(z|y) \notag \\
& + \frac{1}{2} \underset{x \sim p_{\theta}(\cdot | z)}{\E} \log p_{\theta}(z|x) + \underset{\underset{\sim p_{\theta}(\cdot,\cdot |x)}{(y,z)}}{\E} \log p(y,z) \notag \\
& + \underset{\underset{\sim p_{\theta}(\cdot,\cdot |y)}{(x,z)}}{\E} \log p(x,z) + \underset{\underset{\sim p_{\theta}(\cdot,\cdot |z)}{(x,y)}}{\E} \log p(x,y) \notag \\
\label{eq:lowerbound}
\end{align}

This lower-bound contains two types of terms. The \emph{back-translation} terms, e.g., 
\begin{equation} \label{eq:bt}
    \underset{y \sim p_{\theta}(\cdot | x)}{\E} \log p_{\theta}(x|y),
\end{equation} enforce that reciprocal translation models are consistent. The joint terms e.g. ${\E}_{(x,y)\sim p_{\theta}(\cdot,\cdot |z)} \log p(x,y)$ will vanish in our optimization procedure, as explained next.

We use the EM algorithm to maximize Eq.~\ref{eq:lowerbound}. In our setup, the \textbf{E}-step at iteration $t$ amounts to computing the expectations against the conditional distributions evaluated at the current set of parameters $\theta = \theta^{(t)}$. We approximate this by removing the expectations and replacing the random variable with the mode of its distribution i.e. 
$\underset{y \sim p_{\theta^{(t)}}(\cdot | x)}{\E} \log p_{\theta^{(t)}}(x|y) \approx p_{\theta^{(t)}}(x|\hat{y})$ where $\hat{y} = \argmax_{y} p_{\theta^{(t)}}(y | x)$.
In practice, this amounts to running a greedy decoding procedure for the relevant translation models.

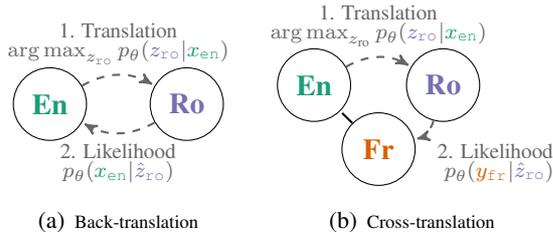
\begin{figure}[t!]
  \centering
  \subfigure[{\scriptsize Back-translation}]{
\begin{tikzpicture}[>=stealth',shorten >=1pt,auto,node distance=1.8cm]
  \node[state] (en) [] {$\color{color1}\textbf{En}$};
  \node[state]         (ro) [right of=en] {$\color{color2}\textbf{Ro}$};

  \path[->, black!60, font=\fontsize{8}{0}\selectfont, dashed, thick]          (en)  edge          [bend left]       node[align=center] {1. Translation \\ $\argmax_{z_{\texttt{ro}}} p_{\theta}({\color{color2}z_{\texttt{ro}}}|{\color{color1}x_{\texttt{en}}})$} (ro);
  \path[->, black!60, font=\fontsize{8}{0}\selectfont, dashed, thick]          (ro)  edge          [bend left]        node[align=center] {2. Likelihood \\ $p_{\theta}({\color{color1}x_{\texttt{en}}} | {\color{color2}\hat{z}_{\texttt{ro}}})$} (en);
\end{tikzpicture}
\label{pic:bt}
  }
  \subfigure[{\scriptsize Cross-translation}]{
\begin{tikzpicture}[>=stealth',shorten >=1pt,auto,node distance=1.2cm]
  \node[state]         (fr) [above left of=en]  {$\color{color1}\textbf{En}$};
  \node[state] (en) [] {$\color{color3}\textbf{Fr}$};
  \node[state]         (ro) [above right of=en] {$\color{color2}\textbf{Ro}$};

  \path[->, black!60, font=\fontsize{8}{0}\selectfont, dashed, thick]          (fr)  edge      [bend left]           node[align=center] {1. Translation \\ $\argmax_{z_{\text{ro}}} p_{\theta}({\color{color2}z_{\texttt{\texttt{ro}}}}|{\color{color1}x_{\texttt{en}}})$} (ro);
  \path[->, black!60, font=\fontsize{8}{0}\selectfont, dashed, thick]          (ro)  edge   [bend left]    node[align=right]  {2. Likelihood  \\$p_{\theta}({\color{color3}y_{\texttt{fr}}} | {\color{color2}\hat{z}_{\texttt{ro}}})$} (en);
  \path[-, thick]          (en)  edge       node {} (fr);
\end{tikzpicture}
\label{pic:ct}
  }
 \caption{Illustration of the back-translation and cross-translation losses. Stop gradient is applied on step 1.}
\end{figure}

The \textbf{M}-step then corresponds to choosing the $\theta$ which maximizes the resulting terms after we perform the \textbf{E}-step. Notice that for this step, the last three terms in Eq.~\ref{eq:lowerbound} no longer possess a $\theta$ dependence, as the expectation was computed in the E-step with a dependence on $\theta^{(t)}$. These terms can therefore be safely ignored, leaving us with only the back-translation terms. By our approximation to the \textbf{E}-step, these expressions become exactly the loss terms that appear in the current UNMT literature \cite{artetxe19, lample19, song19}, see Figure~\ref{pic:bt} for a graphical depiction. Since computing the argmax is a difficult task, we perform a single gradient update for the \textbf{M}-step and define $\theta^{({t+1})}$ inductively this way.

\subsection{Auxiliary parallel data} 
\label{subsec:auxparallel}
We now extend our framework with an auxiliary parallel corpus (Figure~\ref{pic:unmtwaux}). We assume that we wish to translate from $\texttt{X}$ to $\texttt{Z}$, and that we have access to a parallel corpus  $\mathcal{D}_{\textbf{x}, \textbf{y}}$ that maps sentences from $\texttt{X}$ to $\texttt{Y}$. To leverage this source of data, we augment the log-likelihood $\mathcal{L}$ as follows: 
\begin{equation}\label{eq:aug_lik}
    \mathcal{L}_{\text{aug}}(\theta) = \mathcal{L}(\theta) + \underset{\underset{\in \mathcal{D}_{\textbf{x},\textbf{y}}}{(x,y)}}{\sum} \, \log \underset{z \sim Z}{\E} \, p_{\theta}(x,y|z)
\end{equation}
Similar to how we handled the monolingual terms, we can utilize the EM algorithm to obtain an objective amenable to gradient optimization. By using the EM algorithm, we can substitute the distribution of $Z$ in Eq. \ref{eq:aug_lik} with the one given by $p_{\theta}(z|x,y)$. The structural assumption we made in the case of monolingual data still holds: given any variable in the triplet $(X,Y,Z)$, the remaining two are independent. Using this assumption, we can rewrite the distribution $p_{\theta}(z|x,y)$ as either $p_{\theta}(z|x)$ or $p_{\theta}(z|y)$. Since we can decompose $\log p_{\theta}(x,y|z) = \log p_{\theta}(x|z) + \log p_{\theta}(y|z)$, we can leverage both formulations with an argument analogous to the one in \S\ref{subsec:monolingual}:
\begin{align}\label{eq:lowerboundaux}
    \log \underset{z \sim Z}{\E} & p_{\theta}(x,y|z) = \log \underset{z \sim Z}{\E} p_{\theta}(x|z)p_{\theta}(y|z) \notag \\
    &\geq \underset{z \sim p_{\theta}(\cdot|y)}{\E} \log p_{\theta}(x|z) \notag \\ 
    &+ \underset{z \sim p_{\theta}(\cdot|x)}{\E} \log p_{\theta}(y|z) \notag \\ 
    &+ \underset{z \sim p_{\theta}(\cdot|y)}{\E} \log p(z) + \underset{z \sim p_{\theta}(\cdot|x)}{\E} \log p(z)
\end{align}

A key feature of this lower bound is the emergence of the expressions: \begin{equation} \label{eq:ct} 
    \underset{z \sim p_{\theta}(\cdot|y)}{\E} \log p_{\theta}(x|z) \text{ and } \underset{z \sim p_{\theta}(\cdot|x)}{\E} \log p_{\theta}(y|z).
\end{equation} Intuitively, those terms ensure that the models can accurately translate from $\texttt{Y}$ to $\texttt{Z}$, then $\texttt{Z}$ to $\texttt{X}$ (resp. $\texttt{X}$ to $\texttt{Z}$, then $\texttt{Z}$ to $\texttt{Y}$). Because they enforce cross-language pair consistency, we will refer to them as \emph{cross-translation} terms. In contrast, the back-translation terms, e.g., Eq.~\ref{eq:bt}, only enforced monolingual consistency. We provide a graphical depiction of these terms in Figure~\ref{pic:ct}.

As in the case of monolingual data, we optimize the full likelihood with EM.  During the \textbf{E}-step, we approximate the expectation with evaluation of the expectant at the mode of the distribution. As with \S\ref{subsec:monolingual}, the last two terms in Eq. \ref{eq:lowerboundaux} disappear in the \textbf{M}-step. 


\subsection{Connections with supervised and zero shot methods}

So far, we have only discussed multilingual unsupervised neural machine translation setups. We now derive the other configurations of Figure~\ref{configs}, that is, supervised and zero-shot translation, through our framework. 
\paragraph{Supervised translation:} Deriving supervised translation is straightforward. Given the parallel data dataset $\mathcal{D}_{\textbf{x},\textbf{y}}$, we can rewrite the likelihood as:
\begin{equation*}
    \sum_{(x,y) \in \mathcal{D}_{\textbf{x},\textbf{y}}} \log p_{\theta}(x,y) = \underset{\underset{\in \mathcal{D}_{\textbf{x},\textbf{y}}}{(x,y)}}{\sum} \log p_{\theta}(y|x) + \log p(x)
\end{equation*}

where the second term is a language model that does not depend on $\theta$.
\paragraph{Zero-shot translation:} We can also connect the cross-translation term to the zero-shot MT approach from~\citet{al19}. Simplifying their setup, they consider three languages $\texttt{X}, \texttt{Y}$ and $\texttt{Z}$ with parallel data between $\texttt{X}$ and $\texttt{Y}$ as well as $\texttt{X}$ and $\texttt{Z}$. In addition to the usual cross-entropy objective, they also add \emph{agreement} terms i.e. $\E_{z \sim p_{\theta}(\cdot|x)} \log p(z|y)$ and $\E_{z \sim p_{\theta}(\cdot|y)} \log p(z|x)$. We show that these agreement terms are operationally equivalent to the cross-translation terms i.e. Eq. ~\ref{eq:ct}. We first obtain the following equality by a simple application of Bayes' theorem: $$\log p_{\theta}(y | z) = \log p_{\theta}(z|y) + \log p(y) - \log p(z).$$ We then apply the expectation operation $\underset{z \sim p_{\theta}(\cdot|x)}{\E}$ to both sides of this equation. From an optimization perspective, we are only interested in terms involving the learnable parameters so we can dispose of the term involving $\log p(y)$ on the right. Applying the same argument to $\log p_{\theta}(x|z)$, we obtain:
\begin{align*}
     & \underset{z \sim p_{\theta}(\cdot|x)}{\E} \log \, p_{\theta}(y|z) +  \underset{z \sim p_{\theta}(\cdot|y)}{\E} \log p_{\theta}(x|z) \\
    &= \underset{z \sim p_{\theta}(\cdot|x)}{\E} \log p_{\theta}(z|y) + \underset{z \sim p_{\theta}(\cdot|y)}{\E} \log p_{\theta}(z|x) \\
    &- \underset{z \sim p_{\theta}(\cdot|x)}{\E} \log p(z) - \underset{z \sim p_{\theta}(\cdot|y)}{\E} \log p(z) 
\end{align*}

By adding the quantity $\underset{z \sim p_{\theta}(\cdot|x)}{\E} \log p(z) + \underset{z \sim p_{\theta}(\cdot|y)}{\E} \log p(z)$ to both sides of this inequality, the left-hand side becomes the lower bound introduced in the previous subsection, consisting of the cross-translations terms. The right-hand side consists of the agreement terms from \citet{al19}. We tried using this term instead of our cross-translation terms, but found it to be unstable. This could be attributed to the fact that we lack $\texttt{X} \leftrightarrow \texttt{Z}$ parallel data, which is available in the setup of~\citet{al19}.

\section{Training algorithms} 



We now discuss how to train the model end-to-end. We introduce a \emph{pre-training phase} that we run before the EM procedure to initialize the model. Pre-training is known to be crucial for UNMT~\cite{lample19, song19}. We make use of an existing method, MASS, and enrich it with the auxiliary parallel corpus if available. We refer to the  EM algorithm described in \S\ref{sec:M-UNMT} as \emph{fine-tuning} for consistency with the literature.

\subsection{Pre-training}
The aim of the pre-training phase is to produce an intermediate translation model $p_{\theta}$, to be refined during the fine-tuning step. We pre-train the model differently based on the data available to us. For monolingual data, we use the MASS objective \cite{song19}. The MASS objective consists of masking 
 randomly-chosen contiguous segments\footnote{We choose the starting index to be 0 or the total length of the input divided by two with 20\% chance for either scenario otherwise we sample uniformly at random then take the segment starting from this index and replace all tokens with a \texttt{[MASK]} token.} of the input then reconstructing the masked portion. We refer to this operation as \texttt{MASK}. If we have auxiliary parallel data, we use the traditional cross-entropy translation objective. We describe the full procedure in Algorithm~\ref{algo:pretrain}.
\begin{algorithm}[t]
\scriptsize
  \textbf{Input}: Datasets $\mathfrak{D}$ , number of steps $N$ \; 
 \caption{ \textsc{Pre-Training} }
  \begin{algorithmic}[1]\label{algo:pretrain}
    \STATE Initialize $\theta \leftarrow \theta_0$
    \FOR{step in 1, 2, 3, ..., $N$}
      \STATE Choose dataset $D$ at random from $\mathfrak{D}$. 
      \IF{$D$ consists of monolingual data} 
        \STATE Sample batch $x$ from $D$.
        \STATE Masked version of $x$: $x_{M} \leftarrow \texttt{MASK}(x)$
        \STATE MASS Loss: $\text{ml} \leftarrow \log p_{\theta}(x | x_{M})$
        \STATE Update: $\theta \leftarrow \text{optimizer\_update}(\text{ml}, \theta)$
      \ELSIF{$D$ consists of parallel data}
        \STATE Sample batch $(x,y)$ from $D$. 
        \STATE $\text{tl} \leftarrow \log p_{\theta}(y | x) + \log p_{\theta}(x|y)$
        \STATE $\theta \leftarrow \text{optimizer\_update}(\text{tl}, \theta)$
      \ENDIF
    \ENDFOR
  \end{algorithmic}
\end{algorithm}

\subsection{Fine-tuning}
During the fine-tuning phase, we utilize the objectives derived in Section \ref{sec:M-UNMT}.
At each training step we choose a dataset (either monolingual or bilingual), sample a batch, compute the loss, and update the weights. If the corpus is monolingual, we use the back-translation loss i.e. Eq. \ref{eq:bt}. If the corpus is bilingual, we compute the cross-translation terms i.e. Eq. \ref{eq:ct} in both directions and perform one update for each term. We detail the steps in Algorithm~\ref{algo:fine-tune}. 
\begin{algorithm}[t]
\scriptsize
 \caption{ \textsc{Fine-tuning} }
  \textbf{Input}: Datasets $\mathfrak{D}$, languages $\mathfrak{L}$, initialize parameters from pre-training $\theta_0$ \; 
  \begin{algorithmic}[1] \label{algo:fine-tune}
    \STATE Initialize $\theta \leftarrow \theta_0$
    \WHILE{not converged} 
      \FOR{$D$ in $\mathfrak{D}$}
          \IF{$D$ consists of monolingual data} 
            \STATE  $l_{D} \leftarrow $ Language of $D$.
            \STATE  Sample batch $x$ from $D$.
            \FOR{$l$ in $\mathfrak{L}, l \neq l_{D}$}
              \STATE $\hat{y}_l \leftarrow $Decode $p_{\theta}( \hat{y}_l | x).$
              \STATE $\text{bt}_{l_{D},l} \leftarrow \log p_{\theta}(x| \hat{y}_l).$
              \STATE $\theta \leftarrow \textnormal{optimizer\_update}(\text{bt}_{l_D,l}, \theta).$
            \ENDFOR
          \ELSIF{$D$ consists of parallel data}
            \STATE Sample batch $(x,y)$ from $D$.
            \STATE $l_x \leftarrow$ Language of $x$.
            \STATE $l_y \leftarrow$ Language of $y$.
            \FOR{$l$ in $\mathfrak{L}, l \neq l_x, l_y$}
              \STATE $\hat{z}_l \leftarrow $Decode $p_{\theta}( \hat{z}_l | x)$
              \STATE $\text{ct} \leftarrow \log p_{\theta}(y | \hat{z}_l)$
              \STATE $\theta \leftarrow \textnormal{optimizer\_update}(\text{ct}, \theta)$
            \ENDFOR
          \ENDIF
        \ENDFOR
    \ENDWHILE
  \end{algorithmic}
\end{algorithm}
\section{Experiments}


We conduct experiments on the language triplets English-French-Romanian with English-French parallel data, English-Czech-German with English-Czech parallel data and English-Spanish-French with English-Spanish parallel data, with the unsupervised directions chosen solely for the purposes of comparing with previous recent work \cite{lample19, song19, ren19, artetxe19}. 

\subsection{Datasets and preprocessing}
We use the News Crawl datasets from WMT as our sole source of monolingual data for all the languages considered. We used the data from years 2007-2018 for all languages except for Romanian, for which we use years 2015-2018. We ensure the monolingual data is properly labeled by using the fastText language classification tool \cite{joulin2016bag} and keep only the lines of data with the appropriate language classification. For parallel data, we used the UN Corpus \cite{ziemski16} for English-Spanish, the $10^9$ French-English Gigaword corpus\footnote{https://www.statmt.org/wmt10/training-giga-fren.tar} for the English-French and the CzEng 1.7 dataset \cite{czeng16:2016} for English-Czech. We preprocess all text by using the tools from Moses \cite{moses}, and apply the Moses tokenizer to separate the text inputs into tokens. We normalize punctuation, remove non-printing characters, and replace unicode symbols with their non-unicode equivalent. For Romanian, we also use the scripts from Sennrich\footnote{https://github.com/rsennrich/wmt16-scripts} to normalize the scripts and remove diacretics.  For a given language triplet, we select 10 million lines of monolingual data from each language and use SentencePiece \cite{kudo18} to create vocabularies containing 64,000 tokens of each. We then remove lines with more than 100 tokens from the training set. 

\subsection{Model architectures}

We use Transformers~\cite{vaswani17} for our translation models $p_{\theta}$ with a 6-layer encoder and decoder, a hidden size of 1024 and a 4096 feedforward filter size. We share the same encoder for all languages. Following XLM \cite{lample19}, we use language embeddings to differentiate between the languages by adding these embeddings to each token's embedding. Unlike XLM, we only use the language embeddings for the decoder side. We follow the same modification as done in~\citet{song19} and modify the output transformation of each attention head in each transformer block in the decoder to be distinct for each language. Besides these modifications, we share the parameters of the decoder for every language. 

\subsection{Training configuration}

For pre-training, we group the data into batches of 1024 examples each, where each batch consists of either monolingual data of a single language or parallel data, but not both at once. We pad  sequences up to a maximum length of 100 SentencePiece tokens. During pre-training, we used the Adam optimizer \cite{kingma14} with  initial learning rate of $0.0002$ and weight decay parameter of 0.01, as well as 4,000 warmup steps and a linear decay schedule for 1.2 million steps. For fine-tuning, we used Adamax~\cite{kingma14}  with the same learning rate and warmup steps, no weight decay, and trained the models until convergence. We used Google Cloud TPUs for pre-training and 8 NVIDIA V100 GPUs with a batch size of 3,000 tokens per GPU for fine-tuning.


\begin{table*}[!h]
\scriptsize
  \centering
\begin{tabular}[t]{l|l l | l l|l l}
\hline
  & $\En-\Fr$ & $\Fr-\En$ & $\En-\De$ & $\De-\En$ & $\En-\Ro$ & $\Ro-\En$  \\ \hline
 \textbf{Models without auxiliary parallel data} & & & & & &\\ 
 XLM \cite{lample19} & 33.4 & 33.3 & 27.0 & 34.3 & 33.3 & 31.8 \\
 MASS \cite{song19} & 37.5 & 34.9 & 28.3 & 35.2 & 35.2 & 33.1 \\ 
 D2GPo \cite{D2GPO} & 37.9 & 34.9 & 28.4 & 35.6 & 36.3 & 33.4 \\ 
 \citet{artetxe19} &36.2  &33.5 & 26.9 & 34.4 & \quad - & \quad - \\ 
 \citet{ren19} &35.4& 34.9 & 27.7 & 35.6 & 34.9 & 34.1 \\ 
 mBART \cite{liu2020multilingual} & \quad - & \quad - & \textbf{29.8} & 34.0 & 35.0 & 30.5 \\
 \emph{M-UNMT} & 36.3 & 33.50 & 25.5 & 32.3 & 34.87 & 32.1 \\ \hline
 \textbf{Models with auxiliary parallel data} & & & & &  & \\ 
  mBART \cite{liu2020multilingual} & \quad - & \quad - & \quad - & \quad - & \quad - & 33.9 \\
  \citet{bai2020unsupervised} (Concurrent work) & 36.5 & 33.4 & 26.6 & 30.1 & 35.1 & 31.6 \\
  \citet{li2020reference} (Concurrent work) & \quad - & \quad - & \quad - & \quad - & 37.1 & 34.7 \\
 \emph{M-UNMT} (Only Pre-Train) &29.2 &33.8 &18.3 &29.0 &25.3 &32.6 \\ 
 \emph{M-UNMT} (Fine-Tuned) &\textbf{38.3} &\textbf{36.1} & 28.7 &\textbf{36.0} &\textbf{37.4} &\textbf{35.8} \\
 \quad \qquad \quad \emph{detok SacreBLEU} &\emph{36.1} &\emph{35.8} & \emph{28.9} & \emph{35.8} & \quad - & \quad - \\ \hline 
\end{tabular}
  \caption{BLEU scores of various models for UNMT.  M-UNMT refers to our approach.  The $\En-\Fr$/$\Fr-\En$ directions were on \textit{newstest2014}, while the $\En-\Ro$/$\Ro-\En$  and and $\En-\De$/$\De-\En$ directions were on \textit{newstest2016}. To be consistent with previous work, we report tokenized BLEU. However, to aid future reproducibility, we also report sacreBLEU scores. We do not report sacreBLEU scores for Romanian since it is common to include additional prepreprocessing from Sennrich\footnote{https://github.com/rsennrich/wmt16-scripts} (such as removing diacretics) which is not natively supported by sacreBLEU. See \ref{subsec:results} for details. }
  \label{results_table}
\end{table*}




\subsection{Results}\label{subsec:results}

\paragraph{Evaluation}We use tokenized BLEU to measure the performance of our models, using the multi-bleu.pl script from Moses. Recent work \cite{post2018call}  has shown that the choice of tokenizer and preprocessing scheme can impact BLEU scores tremendously. Bearing this in mind, we chose to follow the same evaluation procedures used\footnote{As verified by their public implementations.} by the majority of the baselines that we consider, which involves the use of tokenized BLEU as opposed to the scores given by sacreBLEU. Given the rise of popularity of SacreBLEU \cite{post2018call}, we also include BLEU scores computed from sacreBLEU\footnote{BLEU+case.mixed+lang.xx-xx+numrefs.1
+smooth.exp+test.wmtxx+tok.13a+version.1.4.14.} on the detokenized text for French and German. We exclude Romanian since most works in the literature traditionally use additional tools from Sennrich not used in sacreBLEU.

\paragraph{Baselines} We list our results in Table \ref{results_table}. We also include the results of six strong unsupervised baselines: 
\textbf{(1)} XLM \cite{lample19}, a cross-lingual language model fine-tuned with back-translation;
\textbf{(2)} MASS \cite{song19}, which uses the aforementioned pre-training task with back-translation during fine-tuning;
\textbf{(3)} D2GPo \cite{D2GPO}, which builds on MASS and leverages an additional regularizer by use of a data-dependent Gaussian prior;
\textbf{(4)} The recent work of \citet{artetxe19} which leverages tools from statistical MT as well subword information to enrichen their models; 
\textbf{(5)} the work of \citet{ren19} that explicitly attempts to pre-train for UNMT by building cross-lingual $n$-gram tables and building a new pre-training task based on them;
\textbf{(6)} mBART \cite{liu2020multilingual}, which pre-trains on a variety of language configurations and fine-tunes with traditional on-the-fly back-transaltion. mBART also leverages Czech-English data for the Romanian-English language pair.

Furthermore, we include concurrent work that also uses auxiliary parallel data: \textbf{(8)} The work of \citet{bai2020unsupervised}, which performs pre-training and fine-tuning in one stage and replaces MASS with a denoising autoencoding objective; \textbf{(9)} the work of \citet{li2020reference} which also leverage a cross-translation term and additionally include a knowledge distillation objective. We also include the results of our model after pre-training i.e. no back-translation or cross-translation objective, under the title \emph{M-UNMT (Only Pre-Train)}.

Our models with auxiliary data obtain better scores for almost all translation directions.  Pre-training with the auxiliary data by itself gives competitive results in two of the three $\texttt{X}-\En$ directions. 
Moreover, our approach outperforms all the baselines which also which also leverage auxiliary parallel data. This suggests that our improved performance comes from both our choice of objectives and the additional data.


\section{Ablations}
We perform a series of ablation studies to determine which aspects of our formulation explain the improved performance. 

\paragraph{Impact of the auxiliary data}
We first examine the value provided by the inclusion of the auxiliary data, focusing on the triplet English-French-Romanian. To that end, we study four types of training configurations: 
\textbf{(1)} Our implementation of MASS \cite{song19}, with only English and Romanian data. 
\textbf{(2)} No auxiliary parallel data during pre-training and fine-tuning with only the multi-way back-translation objective
\textbf{(3)} No parallel data during the pre-training phase but available during the fine-tuning phase, allowing us to leverage the cross-translation terms. 
\textbf{(4)} Auxiliary parallel data available during both the pre-training and the fine-tuning phases of training. We also include the numbers reported in the original MASS paper \cite{song19} as well as the best-performing model of the WMT'16 Romanian-English news translation task \cite{sennrich16} and report them in Table \ref{table:aux_table}.

The results show that leveraging the auxiliary data induces superior performance, even surpassing the supervised scores of~\citet{sennrich16}. These gains can manifest in either pre-training or fine-tuning, with superior performance when the auxiliary data is available in both training phases.

\begin{table}
\scriptsize
\centering
\scriptsize
\begin{tabular}{l|l|l}
\hline
 Configuration & $\En-\Ro$ & $\Ro-\En$  \\ \hline
 \textbf{Bilingual configurations} & & \\
 MASS \cite{song19}& 35.20 & 33.10 \\ 
 MASS (Our implementation) & 34.14 & 31.78 \\ \hline
 \textbf{M-UNMT configurations} & & \\ 
 \emph{No auxiliary data}. & 34.87 & 32.10 \\ 
 \emph{Auxiliary data in fine-tuning} & 36.57 & 34.32 \\ 
 \emph{Auxiliary data in both phases} & 37.4 & 35.75 \\ \hline
 \textbf{Supervised} & & \\
 \cite{sennrich16} & 28.2 & 33.9 \\
 mBART \cite{liu2020multilingual} & 38.5 & 39.9 \\ \hline
\end{tabular}
  \caption{$\En-\Ro$ and $\Ro-\En$ BLEU scores on \textit{newstest2016} for different ways of leveraging multilinguality and the auxiliary parallel data. M-UNMT refers to our approach.}
  \label{table:aux_table}
\end{table}

\begin{table}[]
\centering
\small
\begin{tabular}{|l|l|l|}
\hline
 Languages & $\En-\Ro$ & $\Ro-\En$  \\ \hline
 $\En, \Fr, \Ro$ & 37.21 & 35.5 \\ 
 $\En, \Es, \Ro$ & 37.38 & 35.21 \\
 $\En, \Cs, \Ro$ & 36.37 & 34.15 \\ \hline
\end{tabular}
  \caption{$\En-\Ro$ and $\Ro-\En$ BLEU scores for varying choices of auxiliary language on WMT \textit{newstest2016}.}
  \label{language_choice_table}
\end{table}

\begin{figure}[t!]
    \subfigure[$\Ro-\En$ BLEU score]{
    \resizebox{3.5cm}{2cm}{
    \begin{tikzpicture}[every axis/.append style={font=\Large}]
    \begin{axis}[mark=none, xmax=100000,  xlabel=Steps, ylabel=BLEU, grid=major, grid style={dotted,black}, style=ultra thick,
    xtick={0, 50000, 100000}, xticklabels = {0, 50k, 100k}]
      \pgfplotstableread[col sep = comma]{tensorboard_data/MonoBT_backward_BLEU.csv}\MonoBTbackwardBLEU
      \addplot[color=color1, dotted] table[x index = {1}, y index = {2}, mark=none]{\MonoBTbackwardBLEU};
      \pgfplotstableread[col sep = comma]{tensorboard_data/MultiBT_backward_BLEU.csv}\MultiBTbackwardBLEU
      \addplot[color=color2, dashed] table[x index = {1}, y index = {2}, mark=none]{\MultiBTbackwardBLEU};
      \pgfplotstableread[col sep = comma]{tensorboard_data/btct_backward_BLEU.csv}\btctbackwardBLEU
      \addplot[color=color3] table[x index = {1}, y index = {2}, mark=none]{\btctbackwardBLEU};
\end{axis}
\end{tikzpicture}}
  }
    \subfigure[$\En-\Ro$ BLEU score]{
    \resizebox{3.5cm}{2cm}{
    \begin{tikzpicture}[every axis/.append style={font=\Large}]
\begin{axis}[mark=none, xmax=100000,  xlabel=Steps, ylabel=BLEU, grid=major, grid style={dotted,black}, style=ultra thick,
    xtick={0, 50000, 100000}, xticklabels = {0, 50k, 100k}]
      \pgfplotstableread[col sep = comma]{tensorboard_data/MonoBT_forward_BLEU.csv}\MonoBTforwardBLEU
      \addplot[color=color1, dotted] table[x index = {1}, y index = {2}, mark=none]{\MonoBTforwardBLEU};
      \pgfplotstableread[col sep = comma]{tensorboard_data/MultiBT_approximate_forward_BLEU.csv}\MultiBTforwardBLEU
      \addplot[color=color2, dashed] table[x index = {1}, y index = {2}, mark=none]{\MultiBTforwardBLEU};
      \pgfplotstableread[col sep = comma]{tensorboard_data/btct_forward_BLEU.csv}\btctforwardBLEU
      \addplot[color=color3] table[x index = {1}, y index = {2}, mark=none]{\btctforwardBLEU};
\end{axis}
\end{tikzpicture}}
  }

    \subfigure[$\Ro  \righttoleftarrow \En $ back-translation loss]{
\resizebox{3.5cm}{2cm}{
\begin{tikzpicture}[every axis/.append style={font=\Large}]
\begin{axis}[mark=none, xmax=100000, xlabel=Steps, ylabel=back-translation loss, grid=major, grid style={dotted,black}, style=ultra thick, xtick={0, 50000, 100000}, xticklabels = {0, 50k, 100k}]
      \pgfplotstableread[col sep = comma]{tensorboard_data/MonoBT_backtranslation_loss.csv}\MonoBTbacktranslationBLEU
      \addplot[color=color1, dotted] table[x index = {1}, y index = {2}, mark=none]{\MonoBTbacktranslationBLEU};
      \pgfplotstableread[col sep = comma]{tensorboard_data/MultiBT_backtranslation_loss.csv}\MultiBTbacktranslationBLEU
      \addplot[color=color2, dashed] table[x index = {1}, y index = {2}, mark=none]{\MultiBTbacktranslationBLEU};
      \pgfplotstableread[col sep = comma]{tensorboard_data/btct_backtranslation_loss.csv}\btctbacktranslationBLEU
      \addplot[color=color3] table[x index = {1}, y index = {2}, mark=none]{\btctbacktranslationBLEU};
\end{axis}
\end{tikzpicture}}
\label{plot:en_bt}
  }
\subfigure[$\En  \righttoleftarrow \Ro$ back-translation loss]{
\resizebox{3.5cm}{2cm}{
\begin{tikzpicture}[every axis/.append style={font=\Large}]
\begin{axis}[mark=none, xmax=100000, xlabel=Steps, ylabel=back-translation, grid=major, grid style={dotted,black}, style=ultra thick,
    xtick={0, 50000, 100000}, xticklabels = {0, 50k, 100k}]
      \pgfplotstableread[col sep = comma]{tensorboard_data/MonoBT_bwd_backtranslation_loss.csv}\MonoBTbwdbacktranslationBLEU
      \addplot[color=color1, dotted] table[x index = {1}, y index = {2}, mark=none]{\MonoBTbwdbacktranslationBLEU};
      \addlegendentry{BT}
      \pgfplotstableread[col sep = comma]{tensorboard_data/MultiBT_bwd_backtranslation_loss.csv}\MultiBTbwdbacktranslationBLEU
      \addplot[color=color2, dashed] table[x index = {1}, y index = {2}, mark=none]{\MultiBTbwdbacktranslationBLEU};
      \addlegendentry{M-BT}
      \pgfplotstableread[col sep = comma]{tensorboard_data/btct_bwd_backtranslation_loss.csv}\btctbwdbacktranslationBLEU
      \addplot[color=color3] table[x index = {1}, y index = {2}, mark=none]{\btctbwdbacktranslationBLEU};
      \addlegendentry{Full}
\end{axis}
\end{tikzpicture}}
\label{plot:ro_bt}
  }
  \caption{Back-translation losses and BLEU scores for the three configurations on our modified version of the WMT'16 dev set.}
  \label{plot:roen_abla}
\end{figure}
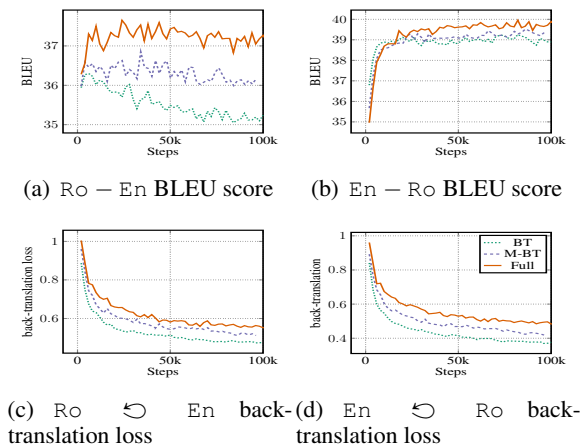

\paragraph{Impact of the additional objectives}

Given the strong performance of our model just after the pre-training phase, it would be plausible that the gains from multilinguality arise exclusively during the pre-training phase. To demonstrate that this is not the case, we investigate three types of fine-tuning configurations: 
\textbf{(1)} Disregard the auxiliary language and fine-tune using only back-translation with English and Romanian data as per~\citet{song19}. \textbf{(2)} Finetune with our multi-way back-translation objective. \textbf{(3)} Finetune with our multi-way back-translation objective and leverage the auxiliary parallel data through the cross-translation terms. We name these configurations \textbf{BT}, \textbf{M-BT}, and \textbf{Full} respectively.
We plot the results of training for 100k steps in Figure \ref{plot:roen_abla}, reporting the numbers on a modified version of the dev set from the WMT'16 Romanian-English competition where all samples with more than 100 tokens were removed.

In the $\Ro-\En$ direction, the  BLEU score of the \textbf{Full} setup dominates the score of the other approaches. Furthermore, the performance of \textbf{BT} decays after a few training steps. In the $\En-\Ro$ direction, the BLEU score for the \textbf{BT} and \textbf{M-BT} reach a plateau about 1 point under \textbf{Full}. Those charts illustrate the positive effect of the cross-translation terms.
We contrast the BLEU curves with the back-translation loss curves in Figure~\ref{plot:en_bt} and \ref{plot:ro_bt}. We see that even that though the \textbf{BT} configuration achieves the lowest back-translation loss, it does not attain the largest BLEU score. This demonstrates that using back-translation for the desired (source, target) pair alone is not the best task for the fine-tuning phase. We see that the multilinguality helps, as adding more back-translation terms with other languages involved improves the BLEU score at the cost of higher back-translation errors. From this viewpoint, the multilinguality acts as a regularizer, as it does for traditional supervised machine translation.  


\paragraph{Impact of the choice of auxiliary language}
In this study, we examine the impact of the choice of auxiliary language. We perform the same pre-training and fine-tuning procedure using either French, Spanish or Czech as the auxiliary language for the English-Romanian pair, with relevant parallel data of this auxiliary language into English. To isolate the effect of the language choice, we fixed the amount of monolingual data of the auxiliary language to  roughly $40$ million examples, as well as roughly $12.5$ million lines of parallel data in the X-English direction.  Table \ref{language_choice_table} shows the results, indicating that using French or Spanish yields similar BLEU scores. Using Czech induces inferior performance, demonstrating that choosing a suitable auxiliary language plays an important role for optimal performance. The configuration using Czech still outperforms the baselines, showing the value of having any auxiliary parallel data at all.



\section{Conclusion and Future Work}

In this work, we explored a simple multilingual approach to UNMT and demonstrated that multilinguality and auxiliary parallel data offer quantifiable gains over strong baselines. We hope to explore massively multilingual unsupervised machine translation in the future.


\bibliography{anthology,emnlp2020}
\bibliographystyle{acl_natbib}

\end{document}

%% file: math_commands.tex
\usepackage{amsmath,amsfonts,bm}

\def\eqref#1{equation~\ref{#1}}

\def\1{\bm{1}}

\def\vx{{\bm{x}}}
\def\vy{{\bm{y}}}

\DeclareMathAlphabet{\mathsfit}{\encodingdefault}{\sfdefault}{m}{sl}
\SetMathAlphabet{\mathsfit}{bold}{\encodingdefault}{\sfdefault}{bx}{n}

\def\gD{{\mathcal{D}}}

\def\gL{{\mathcal{L}}}

\newcommand{\E}{\mathbb{E}}

\DeclareMathOperator*{\argmax}{arg\,max}